\documentclass[letterpaper, 10 pt, conference]{ieeeconf}  % Comment this line out if you need a4paper
\IEEEoverridecommandlockouts
\usepackage{graphics} % for pdf, bitmapped graphics files
\usepackage{epsfig} % for postscript graphics files
\usepackage{mathptmx} % assumes new font selection scheme installed
\usepackage{times} % assumes new font selection scheme installed
\usepackage{amsmath} % assumes amsmath package installed
\usepackage{amssymb}  % assumes amsmath package installed

\usepackage{graphicx}
\usepackage{subcaption}
\usepackage{mwe}
\usepackage{siunitx}

\usepackage{xcolor}

\newcommand{\mbf}[1]{\mathbf{#1}}
\usepackage{amsmath}
\usepackage{flushend}

\usepackage{balance}

\DeclareMathOperator*{\argmin}{arg\,min}

\graphicspath{{./images/}}

\title{\LARGE \bf
Attainment Regions in Feature-Parameter Space for High-Level Debugging in Autonomous Robots
}

\author{Sim\'on C. Smith and Subramanian Ramamoorthy$^{1}$%
\thanks{This work was supported by EPSRC funding for the ORCA Hub
(EP/R026173/1, 2017-2021).}%
\thanks{$^{1}$S. C. Smith and S. Ramamoorthy are with the Institute of Perception, Action and Behaviour, School of Informatics, The University of Edinburgh, 10 Crichton Street, EH8 9AB, UK. SCS is also with the Adaptive \& Intelligent Robotics Lab at Imperial College London.
        {\tt\small \{artificialsimon, s.ramamoorthy\}@ed.ac.uk}}%
}

\begin{document}
\maketitle
\thispagestyle{empty}
\pagestyle{empty}

%%%%%%%%%%%%%%%%%%%%%%%%%%%%%%%%%%%%%%%%%%%%%%%%%%%%%%%%%%%%%%%%%%%%%%%%%%%%%%%%
\begin{abstract}
  Understanding a controller's performance in different scenarios is crucial for robots that are going to be deployed in safety-critical tasks. If we do not have a model of the dynamics of the world, which is often the case in complex domains, we may need to approximate a performance function of the robot based on its interaction with the environment.
  Such a performance function gives us insights into the behaviour of the robot, allowing us to fine-tune the controller with manual interventions.
  In high-dimensionality systems, where the action-state space is large, fine-tuning a controller is non-trivial.
  To overcome this problem, we propose a performance function whose domain is defined by external features and parameters of the controller.
  Attainment regions are defined over such a domain defined by feature-parameter pairs, and serve the purpose of enabling prediction of successful execution of the task.
  The use of the feature-parameter space --in contrast to the action-state space-- allows us to adapt, explain and fine-tune the controller over a simpler (i.e., lower dimensional) space. 
  When the robot successfully executes the task, we use the attainment regions to gain insights into
the limits of the controller, and its robustness. When the robot fails to execute the task, we use the regions to debug the controller and find adaptive and counterfactual changes to the solutions.
Another advantage of this approach is that we can generalise through the use of Gaussian processes regression of the performance function in the high-dimensional space.
 To test our approach, we demonstrate learning an approximation to the performance function in simulation, with a mobile robot traversing different terrain conditions. Then, with a sample-efficient method, we propagate the attainment regions to a physical robot in a similar environment.
  
\end{abstract}

%%%%%%%%%%%%%%%%%%%%%%%%%%%%%%%%%%%%%%%%%%%%%%%%%%%%%%%%%%%%%%%%%%%%%%%%%%%%%%%%
\section{INTRODUCTION}
Many interesting tasks in autonomous robotics include scenarios where the interaction between the robot and the environment involves complex and non-linear dynamics. This complexity makes it hard the have access to an explicit model of the dynamics. One way to cope with the lack of a model is to use model-free learning techniques. In this learning paradigm, control policies are learned directly from the interaction of the robot and the environment. Examples of these techniques include model-free RL (Reinforcement Learning) and supervised learning like LfD (Learning from Demonstration)~\cite{sutton2018reinforcement, schaal1997learning}. In these methods, the robot interacts with the environment by choosing actions through a combination of exploration and exploitation policies, or by being guided by a supervisor.

Another approach is to approximate the dynamics based on data from interaction between the robot and the environment.
Learning this approximation is not trivial as the dynamics may include non-linearities, hidden variables, and intractable spaces. 

In feedback control, a controller chooses actions based on the perceived state of the environment and some internal parameters.
These types of controllers are usually described as a function of the action-state space.
In a model-free paradigm, after the execution of the task, a data point in the action-state space can be mapped to a performance value. An example would be a reward signal that depends on the final position of the robot in the environment.
Based on the performance, the action-state space can be divided into subsets with different properties. These divisions, or regions, have been used to produce hierarchies of \emph{safe workspace regions} where the system remains within the region and converges to the goal~\cite{burridge1999sequential}. Other uses of these regions are viability kernels or backward-reachable sets~\cite{heim2020learnable, bansal2017hamilton}, where an agent can safely explore the action-state space without reaching a failure state. In general, these regions can be used for robust control, explainability, and to fine-tune of the controller.  

Such region analysis on the action-state space can be intractable for high-dimensional systems. Most realistic systems and tasks would indeed involve high-dimensionality of the state-action space. 
Some approaches to learning the performance function in high-dimensional systems involve neural networks (NNs). The drawback of using such NN architectures is that the regression performance comes at the computational cost of requiring large training data sets, and inscrutability of the decisions.
Another approach to learning erformance function is Gaussian process (GP) regression~\cite{heim2020learnable}.  GPs can be used to model the dynamics by making assumptions over the data distribution, but their use can be restricted to lower-dimensional domains.

In this work, we propose to learn the performance of the system based on features of the state and parameters of the controller. We define attainment regions over the domain of the performance function as the set of feature-parameter pairs that produce a successful execution of the task.
A feature-parameter pair represents the initial conditions of a single execution of the task. If the result of the execution of the task is a success, the initial condition pair is part of the attainment region.
Thus, we can identify the attainment regions in a model-free way. 

Compared to the action-state space, the feature-parameter space can represent subsets or regions of the former using a reduced number of variables, hence allowing the use of learning architectures that are not too expensive. Features can represent patterns that are common to a set of states. For example, images belonging to a training set can be encoded as a small number of features obtained in the latent space of a VAE (variational autoencoder~\cite{kingma2013auto}), thus reducing the dimensions of the original state space.
In the case of controller parameters, such as for a Proportional-integral-derivative (PID) controller, the selection of actions can be represented by only three parameters. However, this isn't directly connected to the environment. The use of features and parameters allow us to study the behaviour in a way that connects this to the environment. Another advantage of using features is the ability to interpolate between states realistically. The interpolation allows the generalisation of the controller and the estimation of the performance in a scenario that has not yet been experienced.
Also, debugging can benefit from this simplified representation. For example, a single dimension of the latent space can identify a full region of the original state space. In this case, we show how we can tune a controller trained only with simulated data for a physical system with only a few samples in the real environment.

There are several advantages to modelling the behaviour of the robot with attainment regions. These regions allow us to understand the limits of the operating regime of the robot. A larger region is more robust to external perturbations than a smaller one. The use of features instead of states allows the system to generalise in performance estimation to states that have not been experienced. As the attainment region and its boundaries can be represented graphically, system designers can debug the system. For example, identifying combinations of features and parameters that require a more detailed control policy. As the system assumes that any point outside the attainment region does not result in a successful trial, we can search for the shortest path in parameter-space between the failure point and the attainment region. The controller can use the proposed parameters to change its internal configuration and adapt for a successful trial. Another type of solution in the case of a failed execution is based on the smallest modification of the state to produce a successful trial. These solutions are termed counterfactuals and are usually out of the reach of the robot. For example, a counterfactual can propose a modification of the environment that requires human intervention. In this case, the counterfactual serves as an explanation of a failure case. 

We test our system with a four-wheeled robot with the task of travelling across a ramp at different inclinations and with different contact materials.
The material of the ramp varies between different friction coefficients, and the steepness of the ramp can be regulated at different values. The task is successful if the robot can traverse to the top of the ramp within a limited time; otherwise, it is a failure. We collect synthetic data from a simulator configured to represent the physical environment as close as possible, including friction coefficient for the different materials. Using a Beta-VAE on the RGB images, we can manually select two latent variables that best represent the materials and steepness of the ramp as features of the state. With the simulation data, we can fit a Gaussian process in the reduced feature-parameter space to predict the performance of the robot based on the image and controller gains.

After training the model with synthetic data, we apply the same method to a physical robot. We show that we can fine-tune a mapping between simulated and physical features. By adjusting the attainment regions to include data collected from the physical robot, we show that with a few samples, the model can adapt to the real sensory-motor loop and generalise to previously unseen configurations. 

\subsection{RELATED WORK}

Defining and finding regions where the performance of a robot is reliably known is crucial for safe deployment in complex environments. The notion of controller robustness is related to gain and phase regions where the robot remains stable even when it is externally disturbed~\cite{zhou1998essentials}. With the calculation of disk margins, based on gain and phase regions, and the explicit definition of the model, controllers can be designed and fine-tuned~\cite{chu1999tuning,blight1994practical}. When there is no access to an explicit model, sampling strategies are used to measure safety boundaries~\cite{heim2020learnable,piovan2015reachability}. A popular approach to learning safety boundaries is the viability kernel. These kernels represent regions in the state space where an agent explores actions leading to success in a specific task. By sampling from the state space, the kernels are used in the synthesis of robust controllers, guaranteeing performance and estimating safety properties~\cite{aubin2011viability,liniger2017real}. Another objective of the region analysis is to explain the decision process of the system~\cite{rahwan2019machine}. When the models have many parameters, e.g. with the use of deep neural networks (DNN), system designers have to study them as a black-box. In this case, one again requires sampling~\cite{zhou2020rocus}. To study such high-dimensional systems, the authors in~\cite{lundberg2017unified} use features to build local models to explain the full model. In~\cite{smith2020counterfactual}, the authors exploit the feature space to measure the robustness of robot control using counterfactuals. Other approaches to the study of these regions are based on the parameter space. For example, in~\cite{fan2020parameter}, the parameter space of a controller is partitioned by regions that verify constraints formulated using Signal Temporal Logic. Tuning parameters is common in PID controllers, but usually, the state space is not considered ~\cite{borase2020review}. Similarly, in dynamical systems and probabilistic models, the parameters can be synthesised for the system to arrive at a user-specified region~\cite{dreossi2017sapo}.

In this paper, we propose learning and exploiting attainment regions in feature-parameter space representing the successful completion of a task.

%%%%%%%%%%%%%%%%%%%%%%%%%%%%%%%%%%%%%%%%%%%%%%%%%%%%%%%%%%%%%%%%%%%%%%%%%%%%%%%%
\section{METHOD}
We define a dynamical systems in the action-state space as $\mbf s' = f(\mbf a, \mbf s)$ with action $\mbf a \in A$, state $s \in S$ and a transition function $f$ to the next state $\mbf s'$. Features of the state can be defined as $\mbf z = z(\mbf s)$ with $\mbf z \in Z$ and $z : S \to Z$. Actions are a function of the state and some internal parameters to a controller $C_\theta$, $\mbf a = C_\theta(\mbf s)$, with parameters $\mbf \theta \in \Theta$ and $C_\theta:~S~\to~A$. Here, we propose a performance measure on the feature-parameter space $p: Z \times \Theta \to \{0, 1\}$, indicating the success or the failure of a task.

The intuition here is that both $\mbf z$ and $\mbf \theta$ represent a subset of $S$ and $A$, respectively. For example, a single feature could represent the position of an object in the environment. In a task of obstacle avoidance, the position of the object influences the final result of a trial. In the state space, when the obstacle is perfectly aligned with the robot, it is represented by a single state. If the obstacle is in front of the robot but slightly shifted to one side, this is represented by a different state. The use of features can represent a subset of these states. For example, one variable representing the broad position of the object either in front, to the left or the right of the robot, thus encapsulating several states in a single variable. In the case of actions, the parameters define the outcome of the controller. In a trial of $n$ steps, for a controller with fixed parameters, the roll-out of actions during all the steps are represented by the initial state and the parameters. In stochastic environments, this relationship still holds, but an approximator of $p$ needs to address this characteristic, as we will show next.

As we do not have access to the function $p$, we need to approximate it. GP (Gaussian processes~\cite{rasmussen2003gaussian}) regression can be used as a probabilistic function approximator. For simplicity, we define $\mbf x = (\mbf z, \mbf \theta)$  with $\mbf x \in X := Z \times \Theta$ the feature-parameter space. The performance function is:
\begin{equation} y = p(\mbf x) = \begin{cases}
  1 & \textrm{if the trial is successful} \\
  0 & \textrm{if the trial failed}.
\end{cases}
\label{eq:performance}
\end{equation}
The probabilistic estimate nature of the GP allows us to introduce prior knowledge of the random variables and estimate an uncertainty of the performance measure, in this case coming from the stochasticity of the system. The posterior estimate of $\mbf x$ is normally distributed, including a set of samples $D = \{ (\mbf x_i, y_i), i = 1,\dots,m\}$ pairs. The estimate $\hat{p}$ is conditioned on the samples $\hat{p}_X(\mbf x) | D \sim \mathcal{N}(\mu(\mbf x), \sigma^2(\mbf x))$,
where $\mathcal N$ is the normal distribution, $\mu$ and $\sigma^2$ are the posterior mean and variance given by the covariance function. Using an RBF (Radial Basis Function) kernel and the set $D$, we can fit the parameters of the GP to approximate $p$.
Authors in~\cite{heim2020learnable} actively sample from the plant. This approach is better suited when the sampling cost is high. 
In our case, we are interested in sampling from synthetic data and later apply it to a physical environment. In comparison, the cost of sampling is relatively low in our approach. As the synthetic data could deviate from the data distribution of the physical system, our system can fine-tune the feature-parameter space. We see this property as a debugging tool for complex systems, and we give details about it at the end of this section.

As the approximator $\hat p$ can estimate the probability of success of a trial based on an instance of $\mbf x$, we can define a threshold parameter $\eta_p$ that indicate the minimum value to include $\mbf x$ in the attainment region. We can calculate the probability that a feature-parameter pair $\mbf x$ belongs to the attainable region as:
\begin{equation}
  P[\hat p_X | D > \eta_p].
\end{equation}

With this definition of the attainment region, we can find solutions for feature-parameter instances that do not belong to the region. Without loss of generality and using multi-dimensional Euclidean distance, we can define a constrained minimisation cost on the shortest path between the actual $\mbf x$ feature-parameter space point outside the attainment region and an optimal point $\mbf x^*$ in the attainment region limit.
\begin{equation}
  \mbf x^* = \argmin_{\mbf x' \in X} || \mbf x - \mbf x' ||.
\label{eq:solution}
\end{equation}

In feature space, autoencoders show an \emph{interpolation} property. The interpolation between two instances in the latent space produces a smooth semantic warping in data space~\cite{berthelot2018understanding}. In parameter space, non-linearities are captured as non-continuous attainment regions, e.g. a blob in feature-parameter space where the task fails, surrounded by an area where the task does succeed.
Using a sequential importance sampling and re-sampling with attribution prior algorithm~(\cite{Burke19Explanation, Smith_2020}), we can find the smallest possible change to $\mbf x$ for a new trial to succeed with probability $\eta_p$. 

We can restrict the sampling of possible solutions to either the feature or parameter space. In this case, we have two different families of solutions. First, the solutions on the parameter space belong to an adaptive controller policy. At the beginning of the trial, with $\hat p_X(\mbf x = (\mbf z, \mbf \theta))$, we can estimate the probability of success of the actual configuration. Restricting the search of the optimal value to the parameter space $\mbf \theta$ and fixing the features $\mbf z$, we can find an optimal set of parameters $\theta^*$. Now, the controller can update its internal representation based on $\theta^*$

The second family of solutions is when we search over the feature space and fix the parameters. This solution corresponds to a counterfactual solution~\cite{smith2020counterfactual}. In this case, the solution is to modify the environment that will allow the trial to succeed. Usually, the robot is not able to carry out the counterfactual modification. If the robot were able to apply said modification, it would be part of the original controller. Counterfactuals give an approximation to what an external agent (autonomous or not) should modify in the environment for the robot to succeed. In the example of obstacle avoidance, in a cluttered environment, where there is no solution to avoid them, removing one or more obstacles will allow the robot to find a solution path. This minimal modification, outside the responsibility of the autonomous robot, is the counterfactual solution. Counterfactuals are used for accountability, as causes outside the responsibility of the robot can explain why a trial has failed. Also, counterfactuals are a debugging tool. A single counterfactual in feature space can indicate regions in the state space where the system fails and requires attention.

%%%%%%%%%%%%%%%%%%%%%%%%%%%%%%%%%%%%%%%%%%%%%%%%%%%%%%%%%%%%%%%%%%%%%%%%%%%%%%%%
\section{EXPERIMENTS \& RESULTS}
To evaluate our approach and to show that we can learn the attainment regions in realistic settings, we define a task involving a physical robot with a high-dimensional sensory-motor loop. 
We work with a Clearpath Robotics Husky A200 UGV (unmanned ground vehicle). This robot us equipped with a Multisense S7 sensor. We use the RGB camera as the main sensor. In our experiments, the goal of the robot is to move along a ramp with different materials and steepness. The materials that we vary in the trials are the metal of the ramp itself and synthetic ice. The metallic ramp is designed to reduce slippage, while the synthetic ice has a reduced friction coefficient. We vary the steepness of the ramp between $0^\circ$ and $30^\circ$. Using a PID controller on the angular velocity of the wheels, the task of the robot is to travel to the end of the ramp successfully. In case that the robot does not arrive, the task is a failure. Failure can occur for two main reasons. First, the robot slips to a side given the different friction  from the diverse contact materials placed on top of the ramp. The second failure cause comes from the steepness of the ramp. The ramp angle relative to the ground may be too high for the robot to climb it. 

The three gains of the PID controller represent the parameter vector $\mbf \theta = (K_P, K_I, K_D)$. The state features are represented by the occurrence of synthetic ice $z_1 = \{0, 1\}$ on the ramp and its steepness $z_2 = [0, 30]$  as the inclination in degrees with respect to the horizontal plane. Depending on the materials:
\[ z_1 = \begin{cases}
  1 & \textrm{if there is synthetic ice} \\
  0 & \textrm{if there is no synthetic ice}.
\end{cases}
\]
This function defines the feature vector $\mbf z = (z_1, z_2)$ and the feature-parameter vector as $\mbf x = [\mbf z: \mbf \theta]$.

\begin{figure}
	\vspace{5pt}
        \centering
        \begin{subfigure}[]{0.49\columnwidth}
            \centering
            \includegraphics[height=75pt]{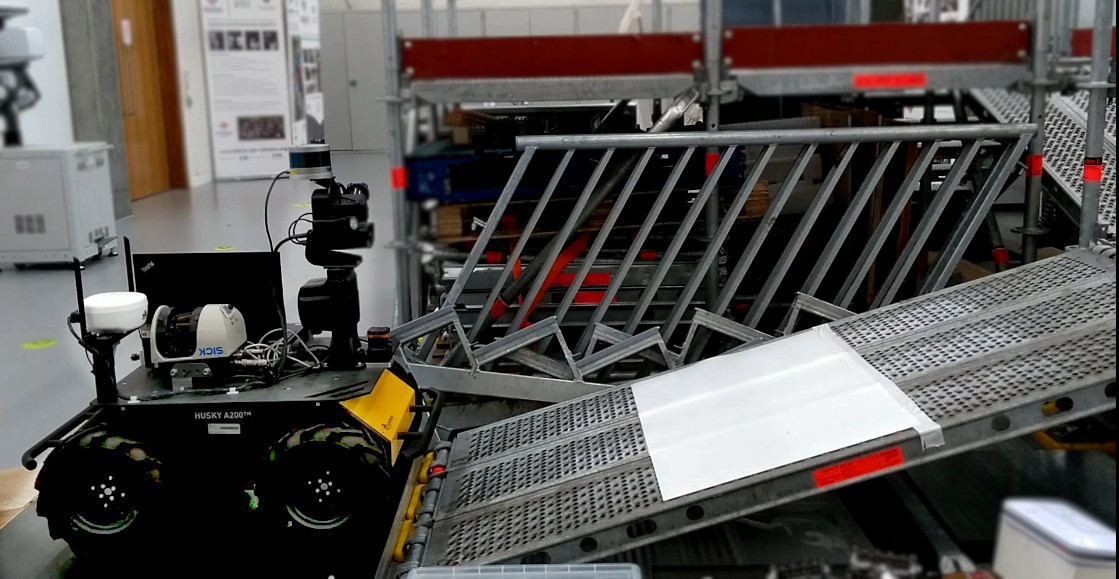}
            \caption[]%
            {{\small Husky}}    
            \label{fig:phys-env}
        \end{subfigure}
        \hfill
        \begin{subfigure}[]{0.49\columnwidth}  
            \centering 
            \includegraphics[height=75pt]{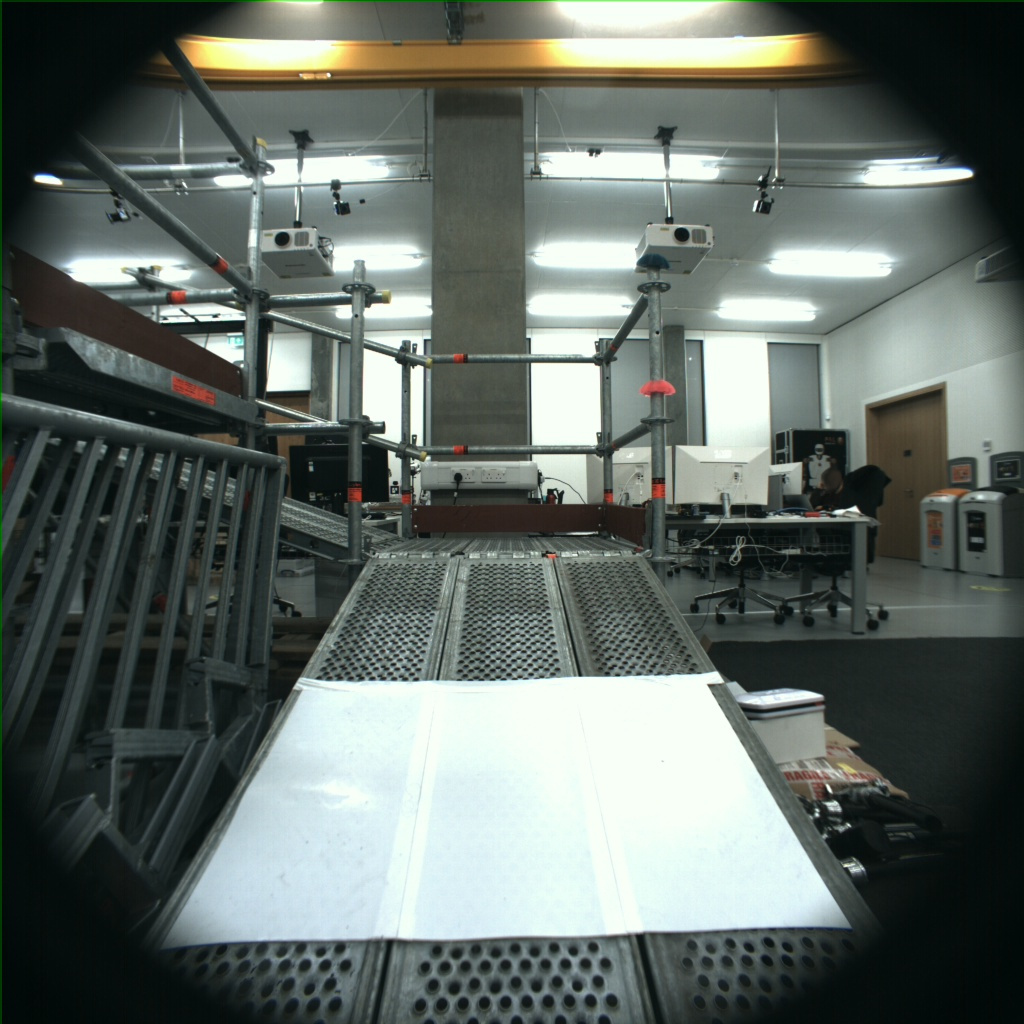}
            \caption[]%
            {{\small Multisense camera}}    
            \label{fig:multisense-camera}
        \end{subfigure}
        \vskip\baselineskip
        %\hfill
        \begin{subfigure}[]{0.49\columnwidth}   
            \centering 
            \includegraphics[height=75pt]{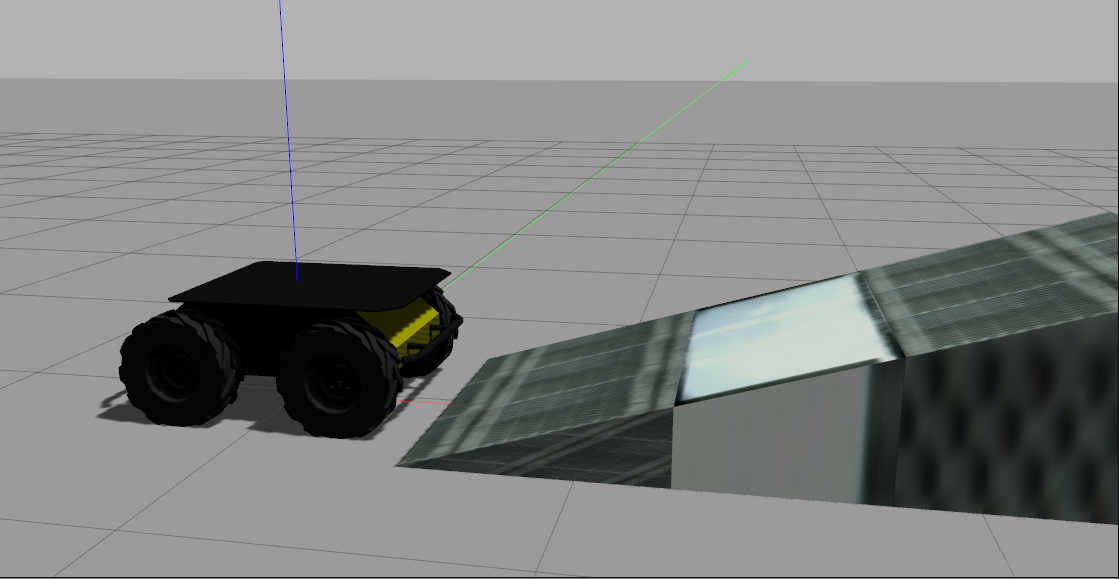}
            \caption[]%
            {{\small Gazebo simulator}}    
            \label{fig:sim-env}
        \end{subfigure}
        \hfill
        \begin{subfigure}[]{0.49\columnwidth}   
            \centering 
            \includegraphics[height=75pt]{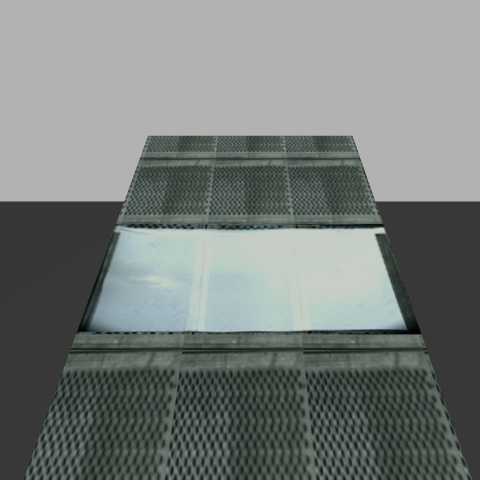}
            \caption[]%
            {{\small Gazebo camera}}    
            \label{fig:gazebo-camera}
        \end{subfigure}
        \caption[]
        {\small (a) and (b) Physical environment and on-board camera. Figs.~(c) and (d) Gazebo simulator environment and RGB camera.} 
        \label{fig:environments}
    \end{figure}

Fig.~\ref{fig:environments} shows the setup of the physical and simulated environment. We use the Gazebo simulator to sample the data points to fit the Gaussian Process estimator $\hat p_X$.
We define both the set-point of the PID controller to a $40\%$ of the maximum angular speed of the wheels and a sampling set for the PID gains. Combining values for $\mbf z$ and $\mbf \theta$, we sample more than 400 points in the feature-parameter space. We run a trial on the simulator for each point and collect the performance result $y$ (Eq.~\ref{eq:performance}). The fitting of the GP was done on a standard desktop computer and took around 1 minute to process all the samples. Including more data points increased the fitting time exponentially. 

To make the simulation result as aligned as possible to the physical environment, we estimate the friction coefficients between the synthetic ice and the rubber wheels of the robot. We sample a few trials in the physical environment and search for the best coefficient to reproduce the same performance results. In our experiment, we can easily estimate the coefficient as we only have two materials. For more complex scenarios, the authors in~\cite{brandao2016friction}  propose a method based on annotated data to estimate the coefficient from pixel-based signals.

\begin{figure}
	\vspace{5pt}
        \centering
        \begin{subfigure}[b]{0.49\columnwidth}
            \centering
            \includegraphics[height=105pt]{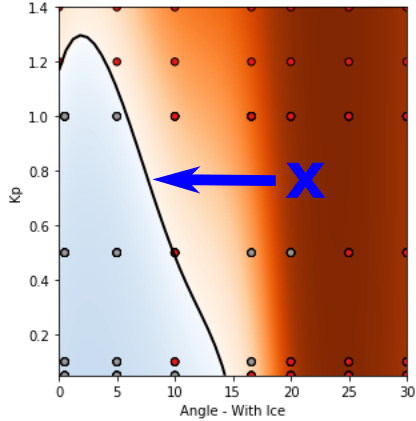}
            \caption[]%
            {$K_P$ vs angle (ice)}    
            \label{fig:ice-kp}
        \end{subfigure}
        \hfill
        \begin{subfigure}[b]{0.49\columnwidth}
            \centering
            \includegraphics[height=105pt]{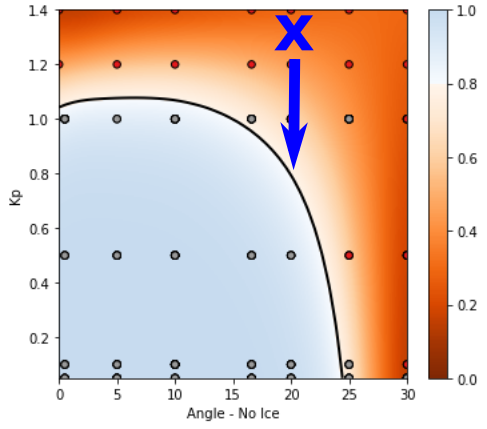}
            \caption[]%
            {$K_P$ vs angle (no ice)}
            \label{fig:noice-kp3}
        \end{subfigure}
        \vskip\baselineskip
        \begin{subfigure}[b]{0.49\columnwidth}
            \centering
            \includegraphics[height=105pt]{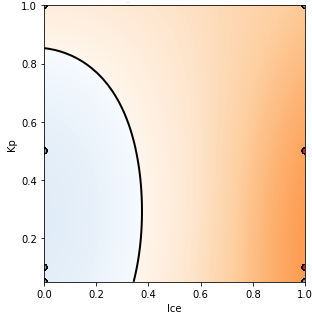}
            \caption[]%
            {$K_P$ vs ice}    
            \label{fig:kp-ice-all}
        \end{subfigure}
        \hfill
        \begin{subfigure}[b]{0.49\columnwidth}
            \centering
            \includegraphics[height=105pt]{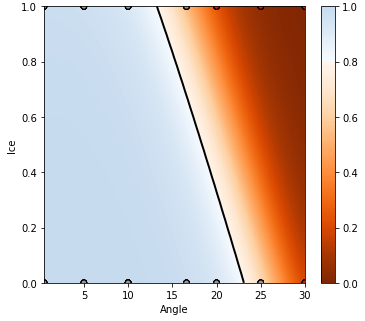}
            \caption[]%
            {Ice vs angle}    
            \label{fig:angle-ice}
        \end{subfigure}
        \vskip\baselineskip
        \begin{subfigure}[b]{0.49\columnwidth}
            \centering
            \includegraphics[height=105pt]{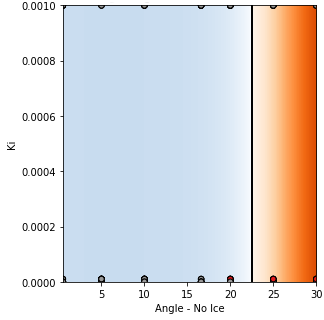}
            \caption[]%
            {$K_I$ vs angle (no ice)}    
            \label{fig:noice-ki}
        \end{subfigure}
        \hfill
        \begin{subfigure}[b]{0.49\columnwidth}
            \centering
            \includegraphics[height=105pt]{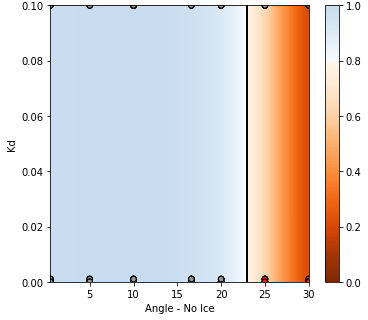}
            \caption[]%
            {$K_D$ vs angle (no ice)}    
            \label{fig:noice-kd}
        \end{subfigure}
        \caption[]
        {\small Light-blue areas represent attainment regions. The blue arrows represent the direction of a counterfactual solution (a) and an adaptive one (b). The dots represent data points. Red dots are failures, and grey ones represent successful trials. Note that at each data point, there may be overlaps of different runs. The images are slices of the 5-dimensional parameter-feature space. In (a), (b), (c) and (d), the variables not shown are not restricted. In (e) and (f), $K_p$ has been restricted to $1.0$.}
        \label{fig:regions}
    \end{figure}

Using an RBF  as the kernel for the Gaussian process approximator, we fit $\hat p_X$ with the results of the simulated trials. By setting a confidence threshold as $\eta_p = 0.8$, we can find the attainment regions of the feature-parameter space where the model will predict an $80\%$ of success rate. Fig.~\ref{fig:ice-kp} shows the attainment region in light-blue, where the robot is expected to succeed with a probability $P(y|\mbf x) \geq 0.8$. In the same subfigure, the horizontal axis represents the angle of the ramp for trials where we used synthetic ice. The vertical axis represents the proportional gain $K_P$ of the PID controller. Fig.~\ref{fig:noice-kp3} shows the attainment region when no synthetic ice is present. Comparing Figs.~\ref{fig:ice-kp}~and~\ref{fig:noice-kp3}, we can see that the occurrence of the ice material reduces the options in the feature-parameters space where the robot can reach the goal. Fig.~\ref{fig:kp-ice-all} shows that a large $K_P$ (vertical axis) can be detrimental when the low-friction material is present. The attainment region in Fig.~\ref{fig:angle-ice} shows that the success of the goal depends on the occurrence of the ice and the steepness of the ramp. Figs.~\ref{fig:noice-ki}~and~\ref{fig:noice-kd} show that the results of the task do not depend on $K_I$ and $K_D$.

\subsection{Physical Robot Implementation}
As we do not have direct access to the features in a physical setup, we need to extract them from sensors attached to the robot. From the sensory data, we can use automatic feature extraction methods.
However, sampling directly from a robot can be expensive.
A standard way to circumvent this limitation is to use a  physical simulator and  tune the parameters to correlate the simulation results with the real ones.
Here, we propose to train a Beta-VAE~\cite{higgins2016beta} to encode and decode images from an RGB camera in the simulated environment and use its latent space as the feature space.
As the autoencoder will be trained on synthetic data, we will use a mapping to correct any deviation in the encoding when using real images from the camera of the physical robot $\mbf z = m(z'(\mbf s))$,
with $z': S \to Z$ as the Beta-VAE encoding.

As we can linearly interpolate the features in an autoencoded latent space, we use a linear function for $m$. A linear model allows us to fit its parameter with just a couple of samples. In our experiments, this model effectively translated the features from the simulator to the physical ones. Note that this mapping allows us to carry the posterior learned during training with the simulated system. Now, the performance predictor in the physical system is:
\begin{equation}
  \hat{p}_X(\mbf x) = \hat{p}_X((\mbf z, \mbf \theta)) = \hat{p}_X((m(z'(\mbf s)), \mbf \theta)).
  \label{eq:performance_approximator}
\end{equation}

\begin{figure}
	\vspace{5pt}
        \centering
         \includegraphics[width=\columnwidth]{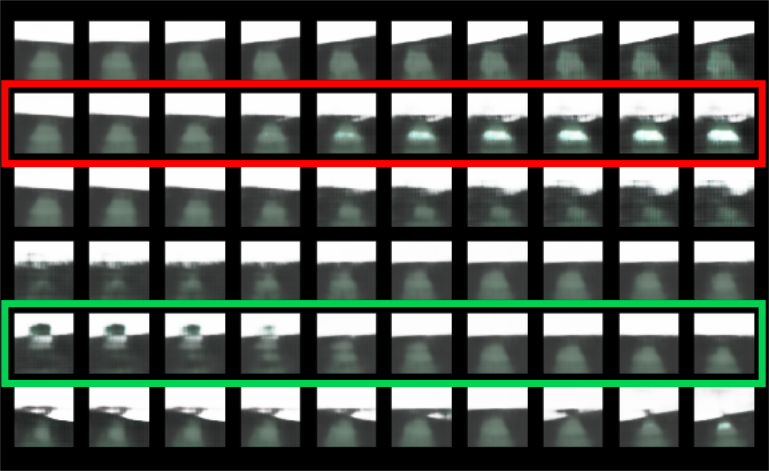}
         \caption[]%
         {\small Latent space of the Beta-VAE trained with images from the onboard camera. Two latent dimensions are selected for the synthetic ice occurrence (red-box) and the ramp steepness (green-box).}     
         \label{fig:features-sel}
\end{figure}

 From the data collected during the simulated trials from the previous step and using data augmentation techniques, we train a Beta-VAE with the RGB images from the onboard camera as a reconstruction target. The meta-parameters of the autoencoder include a latent dimension of size 32, $\beta = 20$, 5 hidden layers of convolutional filters with batch normalisation, ReLU activation function and a dropout rate of $0.5$ in both the encoder and decoder model. Fig.~\ref{fig:features-sel} shows a sample of manipulation of the latent space. Each row is a linear modification of a latent dimension. To map the latent dimensions to the features used to train the Gaussian process, we select two latent variables that best represent the occurrence of synthetic ice and the steepness of the ramp. In Fig.~\ref{fig:features-sel}, the row marked with red shows the initial decoding of the latent dimension where no white blob can be seen, while it gradually appears when the latent variable value is increased. The white blob in the image indicates whether synthetic ice is present or not. In the same figure, the green box indicates a latent variable where the height of the ramp decreases. We associate the height of the ramp with its steepness feature. Note that the goal of the autoencoder is to extract the features, and it does not aim at having a perfect reconstruction. In this experimental setup, we have used expert knowledge to choose the features. In more general cases, automatic selection of the features can achieve similar results at the expense of explainability. For example, the system could use all the latent variables in the latent space of the autoencoder. In this case, our method would still find the attainment regions and the solutions in parameter and feature space. However, counterfactual solutions in feature space could be harder to understand by the user. The features may not have a clear representation in the environment or may present  a correlation in the data that does not necessarily have a causal relation with the task, e.g. the sharpness of the RGB image.

To fit the linear model, we select four images representing the maximum and minimum steepness and the occurrence and absence of synthetic ice in the physical environment. Encoding each image, we obtain values in the latent space: for the lowest angle, the value is $0.095$, for the highest angle is $-1.63$, for the occurrence of ice is $1.26$, and for the absence of it, a value of $0.35$. With two points for each latent dimension, we can calculate the linear mapping:
\begin{align}
  z_1 = m_I(x) &= 1.10x - 0.38, 
  \label{eq:linear_map_ice} \\
  z_2 = m_A(x) &= -17.39x + 1.65,
  \label{eq:linear_map}
\end{align}
with $m_A$ and $m_I$ the functions for angle and the occurrence of synthetic ice, respectively. 

We use the attainment regions to find solutions on the physical Husky when the prediction is `failure'. Using the known PID parameters and extracting the state features from the camera, we can estimate the outcome of a trial. If the value is outside the attainment region, we can find a solution either on the feature or parameter space. The simplest solution is to find the smallest modification to $\mbf \theta$ that can move the performance prediction inside the attainment region with fix features variables. To find this modification, we use the sampling optimisation process~\cite{Burke19Explanation} constrained to the prior ($\hat p_X$) and the minimum Euclidean distance to the attainment region. This solution correspond to moving on the parameter space, e.g. horizontal axis on Fig.~\ref{fig:noice-kp3} blue $\mbf x$, from a value of $K_P = 1.3$ to a value in the region limit of $K_P = 0.8$. The controller can adapt to these new parameters online and have a successful trial with parameters that have not been directly experienced during training. Using the image taken from the onboard camera, Fig.~\ref{fig:multisense-camera}, we extract the latent values and map them with Eqs.~\ref{eq:linear_map_ice} and~\ref{eq:linear_map} to the feature space. We define the parameters of the controller as the feature-parameter vector $\mbf x = [z_1, z_2, K_P, K_I, K_D] = [1.09, 13.53, 0.05, \num{1e-6}, 0]$ that fails the task. The prediction of $\hat p_X(\mbf x)$ is $0.52$, outside of the attainment region. Freezing $z_1$ and $z_2$, the system find the new optimal parameter $\mbf \theta^* = [0.5, 0.00001, 0.0]$ with performance  of $0.802$, in the limit of the attainment region. We test these parameters on the Husky on three trials. All the trials succeed in the task.

The other type of solution that this system can provide are counterfactuals. Fig.~\ref{fig:ice-kp} blue $\mbf x$ point shows that in an angle around $22^\circ$ with ice on the ramp, the Husky will fail to meet the goal.
 The solution, moving on the feature space, shows a counterfactual where reducing the angle of the ramp to $8^\circ$ would produce a successful trial with the required minimum performance. We can also fix the angle of the ramp, so the system is forced to find a solution in the rest of the feature, i.e. the synthetic ice. In this case, the solution is at $z_1^* = 0.53$, that using a threshold in the middle point of the linear mapping for Eq.~\ref{eq:linear_map_ice}, represents the absence of ice as another counterfactual solution.

%%%%%%%%%%%%%%%%%%%%%%%%%%%%%%%%%%%%%%%%%%%%%%%%%%%%%%%%%%%%%%%%%%%%%%%%%%%%%%%%
\section{CONCLUSION}

We introduce the concept of attainment regions. Through our experiments, we have shown that these regions can be used to achieve generalisation, robustness, explainability, adaptability and debugging of an autonomous mobile robot in realistic settings. These characteristics are relevant for trustworthy autonomous robots in critical tasks. The advantage of our definition of attainment regions compared to analysis of performance in action-state space comes from its compact representation based on features and parameters.
We demonstrated extraction of features from an image-based sensory input using autoencoders. More advanced feature extraction techniques would facilitate even more sophisticated features within the proposed methodology.

%%%%%%%%%%%%%%%%%%%%%%%%%%%%%%%%%%%%%%%%%%%%%%%%%%%%%%%%%%%%%%%%%%%%%%%%%%%%%%%%
%\balance
\bibliographystyle{IEEEtran}
\bibliography{IEEEabrv,root}

\end{document}